\pdfoutput=1

\documentclass[11pt]{article}

\usepackage{ACL2024}

\usepackage{times}
\usepackage{latexsym}

\usepackage[T1]{fontenc}

\usepackage[utf8]{inputenc}

\usepackage{microtype}

\usepackage{inconsolata}
\usepackage{xcolor}
\usepackage{multirow}
\usepackage{booktabs}
\usepackage{makecell}
\usepackage{amssymb}
\usepackage{colortbl}
\usepackage{graphicx}
\usepackage{bbding}
\usepackage{utfsym}
\usepackage{caption}
\usepackage{subcaption}
\usepackage{amsmath}
\usepackage{enumitem}

\usepackage[export]{adjustbox}

\definecolor{ao(english)}{rgb}{0.0, 0.5, 0.0}
\usepackage{pifont}
\usepackage{xcolor}
\usepackage{xspace}
\definecolor{ForestGreen}{RGB}{34,139,34}
\definecolor{BrickRed}{rgb}{.72,0,0}
\definecolor{LakeBlue}{RGB}{0,61,153}

\title{PathReasoner: Modeling Reasoning Path with Equivalent Extension \\for Logical Question Answering}

\author{Fangzhi Xu$^{1,2}$, Qika Lin$^{1,2}$\thanks{\, Correspondence to Qika Lin and Jun Liu.}, Tianzhe Zhao$^{1,3}$, Jiawei Han$^{1,3}$, Jun Liu$^{1,2*}$ \\
        $^1$School of Computer Science and Technology, Xi’an Jiaotong University \\
        $^2$Ministry of Education Key Laboratory of Intelligent Networks and Network Security \\
        $^3$Shaanxi Province Key Laboratory of Big Data Knowledge Engineering \\
        \texttt{fangzhixu98@gmail.com,qikalin@foxmail.com,liukeen@xjtu.edu.cn}
        }

\begin{document}

\maketitle

\begin{abstract}
Logical reasoning task has attracted great interest since it was proposed. Faced with such a task, current competitive models, even large language models (e.g., ChatGPT and PaLM 2), still perform badly. Previous promising LMs struggle in logical consistency modeling and logical structure perception.
 To this end, we model the logical reasoning task by transforming each logical sample into reasoning paths and propose an architecture \textbf{PathReasoner}. It addresses the task from the views of both data and model. To expand the diversity of the logical samples, we propose 
	an atom extension strategy supported by equivalent logical formulas, to form 
	new reasoning paths. From the model perspective, we design a stack of 
	transformer-style blocks. In particular, we propose a path-attention module to 
	joint model in-atom and cross-atom relations with the high-order diffusion 
	strategy. Experiments show that PathReasoner achieves competitive 
	performances on two logical reasoning benchmarks and great generalization abilities.
\end{abstract}

\section{Introduction}
With the emergence of pre-trained language models (PLMs) \cite{kenton2019bert, brown2020language}, recent years have witnessed remarkable progress in the task of machine reading comprehension (MRC) \cite{rajpurkar2016squad, lai2017race}. To tackle more complex scenarios in reality, the challenging logical reasoning 
task \cite{yu2019reclor, liu2021logiqa} has been proposed to exploit the model reasoning capability \cite{DBLP:conf/acl/0009C23} over text\footnote{Logical reasoning is a broad concept covering various tasks, but we mainly address the task in the form of MRC.}. Similar to the traditional MRC task, it also takes the context, question and options as inputs and requires the model to predict the final answer. Due to the diverse logical characteristics implied in the text, logical reasoning task brings huge challenges to current LMs. Especially, faced with such tasks, large language models (LLMs), e.g., ChatGPT\footnote{https://chat.openai.com} and PaLM 2\footnote{https://ai.google/discover/palm2/}, also struggle a lot which is proved by previous evaluation works \cite{xu2023large,liu2023evaluating}. Under such circumstances, this paper will focus more on addressing logical reasoning tasks with small LMs, which are light-weighted and more flexible for future applications\footnote{The focus of this paper is mainly on small LMs, since they are more efficient and effective compared with LLMs on the logical reasoning tasks. But we still report LLM performances for comparison in the experiment section.}.

\begin{figure}[t]
	\large
	\centering
	\includegraphics[scale=0.48]{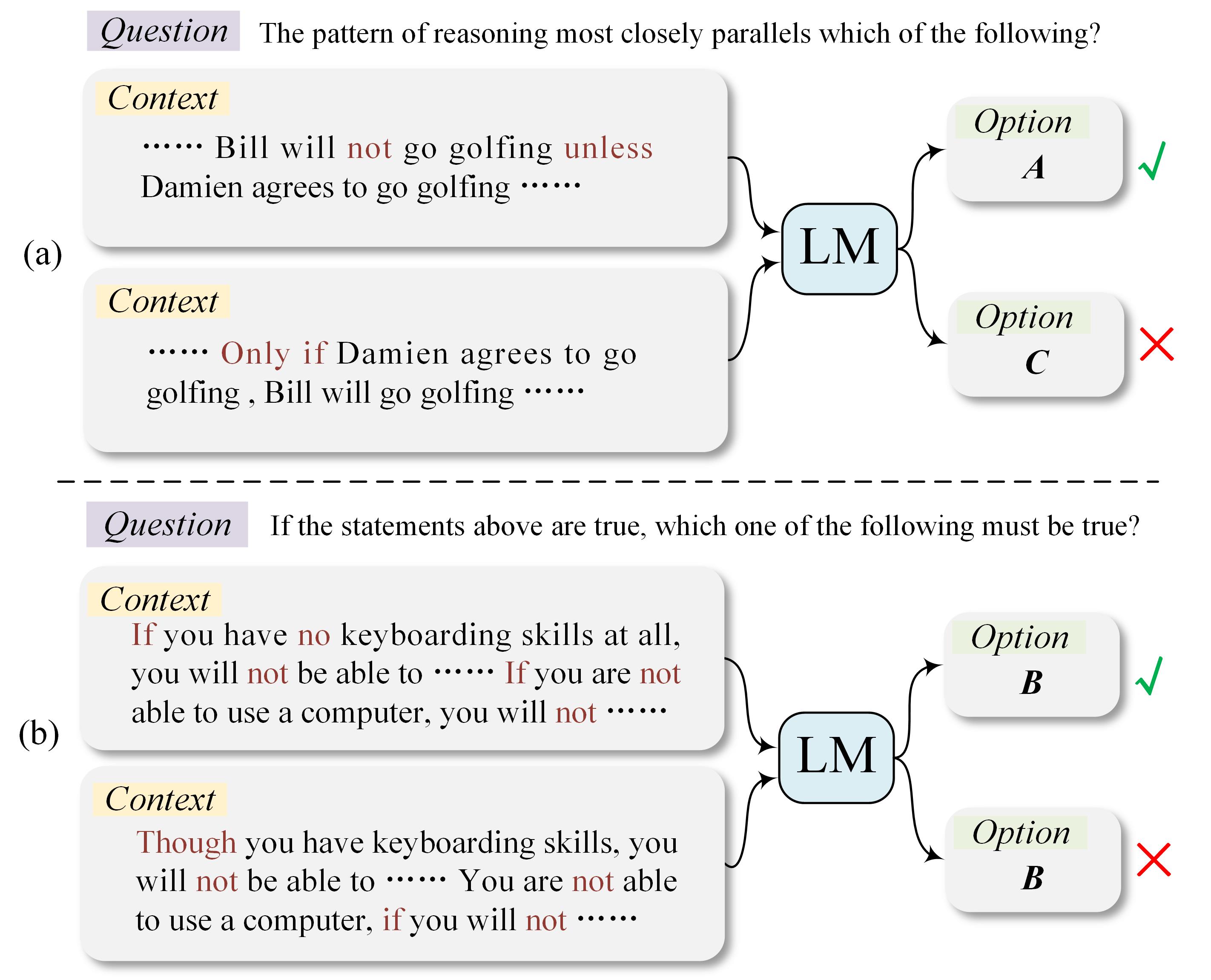}
	\caption{Probing tests on representative LMs (e.g., RoBERTa). (a) is about model prediction consistency. (b) is related to the perception of logical connectives. Detailed pilot experiments are shown in the Appendix.}
	\label{bad_case}
\end{figure}

Previous competitive LMs expose two limitations in the performance. Firstly, it lacks consistent model predictions on samples with equal logical semantics. For example in Figure \ref{bad_case}(a), we make changes to the expression of the original context while maintaining the semantic unchanged, where \emph{not...unless} is equally transformed into the expression of 
\emph{only if}. However, the LMs give inconsistent predictions between the 
original sample and the modified one. We blame the problem on the lack of 
training samples in logical reasoning. Compared with some classic MRC datasets 
like SQuAD \cite{rajpurkar2016squad, rajpurkar2018know}, CoQA 
\cite{reddy2019coqa} with over 100,000 training samples, logical reasoning 
datasets like ReClor \cite{yu2019reclor} and LogiQA~\cite{liu2021logiqa} are 
much more sparse with only several thousand samples. Thus, such sparsity limits 
the learning of logic semantics. Previous work~\cite{jiao2022merit} 
leverages general corpus to conduct continual pretraining, but it does not 
address the sparsity of logical text in essential.

Secondly, it remains a challenge to enhance the model perception for logical 
structures. For example in Figure \ref{bad_case}(b), we randomly replace the 
explicit logical relation words or inverse the negations for the context, which 
destroys the original semantics. But the LMs fail to change the prediction 
accordingly. It demonstrates that current LMs are insensitive to the logical 
connectives, instead they focus more on facts within the text. Considering that 
current LMs are pre-trained with general objectives on the fact corpus (e.g., Wikipedia), they are naturally weak 
in capturing the logical structures usually existing in logical-specific 
scenarios. Some studies like DAGN \cite{huang2021dagn}, Logiformer 
\cite{xu2022logiformer}, and AdaLoGN \cite{li2022adalogn} have attempted to 
model the explicit logical relations from various perspectives, such as causal 
and 
co-occurrence. All of them build text graphs to conduct the reasoning, 
which limits the scalability to larger text and more complex 
scenarios.

In view of the above challenges, we propose an architecture
\textbf{PathReasoner}, which considers a new paradigm for logical reasoning 
tasks via reasoning 
path modeling. Based on the predefined logical rule forms, we represent each 
natural sentence as an atom and transform each sample into reasoning paths with 
confidence scores. Under such a paradigm, PathReasoner addresses the task from 
two views. From the view of expanding the data 
diversity, we first 
obtain equivalent atom combinations through external logical formulas, 
generating new reasoning paths and textualizing them as new samples. From the 
model view, we propose a reasoning 
path modeling network. It encodes both function symbols and variables in atoms and forms an atom embedding sequence as the input. In a path-attention module, we model high-order relations from both in-atom and cross-atom perspectives. Through the fusion of token, atom, and path embedding, the prediction can be derived. 

Our technical contributions are as follows, and additional key values are in Appendix~\ref{appendix:key_novelty}:

\noindent (1) We unify the text inputs into atoms and reasoning paths. Based on it, an 
architecture PathReasoner is proposed to improve both the diversity of samples 
and logic perception capability.

\noindent (2) In light of the sparsity of training data, we propose an atom extension strategy to form new training samples. To better capture logical structures, we introduce a path-attention module with high-order relation modeling, enabling joint updates of information within atoms and across atoms.

\noindent (3) Extensive experiments show superior performances on two 
logical reasoning benchmarks. Significant generalization capabilities are also verified.

\section{Related Work}
Recent progress in MRC promotes the emergence of more complex tasks like logical reasoning. Previously, several datasets on logical reasoning have been proposed, including ReClor \cite{yu2019reclor}, LogiQA \cite{liu2021logiqa} and AR-LSAT~\cite{zhong2021ar}. 
They have attracted much attention since some LMs fail to show superiority. Previous works on the logical reasoning task can be categorized into two folds.

\paragraph{Sequence-based.} These models are usually accompanied by data augmentation 
strategies. LReasoner \cite{wang2022logic} proposes to extend 
text with logical formulas to enrich the context information. MERIt 
\cite{jiao2022merit} proposes a contrastive strategy based on the meta-path 
and leverages the extra data to pre-train the model. However, both of them lack 
the relation modeling of logical units in the sequence.

\paragraph{Graph-based.} DAGN \cite{huang2021dagn} is the first work to divide 
the text into discourse units and utilize the graph neural networks \cite{zhou2020graph} to update 
the representations. But its chain-type graph structure limits the expression of 
complex relations between logical units. FocalReasoner \cite{ouyang2021fact} focuses on 
the fact triplet extracted from the text and builds a supergraph for reasoning. 
But it ignores the effects of the logical connectives within the text. To 
better model the logic within text, AdaLoGN \cite{li2022adalogn} designs an 
adaptive network to update the text graph progressively. Logiformer 
\cite{xu2022logiformer} proposes a two-branch graph transformer network to 
address the text from syntax and logic. However, it is costly 
to form and update the text graph during the reasoning process. In general, the 
graph-based methods naturally lack expansibility, especially when the text 
becomes larger.

Considering the above drawbacks, we propose a reasoning pattern based on the 
reasoning paths (instantiated logical rules) for the first time. It models the 
logical reasoning task from a special perspective and 
combines the advantages of both sequence and graph-based methods.

\section{Preliminary}
\label{preliminary}
This work considers unifying the inputs into the form 
of logical rules~\cite{lin2021contrastive} since it is a more natural way to uncover logical structures of the text while maintaining the important facts.
The distinctive values of such definitions over first-order logic~\cite{xu2023symbol} and propositional logic are in Appendix~\ref{appendix:distinction_logic}.
We introduce the following two definitions.

\noindent \textbf{Definition 1: atom.} We 
transform each natural sentence into one atom~\cite{hinman2018fundamentals}, which consists of one function 
symbol and several variables. For example, given 
the sentence \emph{Paula will visit the dentist only if Bill goes golfing}, we 
define the expression $\texttt{OnlyIf}(A,B)$ as the atom. \texttt{OnlyIf} is 
the function symbol that denotes the 
explicit connective phrase in the sentence. And $A, B$ are called variables to 
represent abstract sentence constitutes, 
whose instantiation are \emph{Paula will visit the dentist} and \emph{Bill goes 
	golfing} respectively. Similarly, we can also derive other atoms from the 
text, 
such as $\texttt{Unless}(A,B)$, $\texttt{Since}(A,B)$, $\texttt{InFact}(A)$.

According to the reasoning patterns, we define four categories 
of function symbols, shown in Table \ref{functions}. The first is causal 
relations for deterministic facts. The second and third ones are conditional 
assumptions, where \emph{NA} focuses on the uniqueness of the condition. The 
last one is facts with no explicit logical relations.

\noindent \textbf{Definition 2: reasoning path.} Based on Definition 1, we can 
unify the context, question and options of each input into the form of the 
logical 
rule~\cite{DBLP:conf/sigir/LinLXPZZZ22,DBLP:conf/emnlp/PanLZZLHW22}, such as Eq. \ref{equation1}:

\begin{equation}
\label{equation1}
\small
\varepsilon,\;\underbrace{F_1(A,B)\land F_2(C,A)\land F_3(D)\land \cdots  
}_{\emph{rule body}} \Rightarrow 
\underbrace{Q(a_i)}_{\emph{rule head}}.
\end{equation}
Rule body functions as the modeling of the context part, which is represented 
as the conjunction of atoms. Rule head consists of the concatenation of 
the question sentence and option $a_{i}$, which is also represented as the 
conjunction of atoms in the implementation. The symbol $\varepsilon$ indicates 
the confidence score of the logical rule. Since each option is bounded with one 
logical rule, $\varepsilon$ is also equal to the confidence of option $a_{i}$. 
In actual cases, the function symbols (e.g., $F_{1}, F_{2}$) and variables 
(e.g., $A,B$) are instantiated as the natural language. Therefore, this paper 
defines the instantiated logical rule as the reasoning path.

\begin{table}[t]
	\centering
	\small
	\begin{tabular}{c|c}
		\toprule
		\textbf{Category} & \textbf{Representative Connectives} \\
		\hline
		Cause &  Because, Since, DueTo, TheReasonsWhy... \\
		SA & If, When, Once, AsLongAs, ... \\
		NA & OnlyIf, Unless, ... \\
		Fact & InFact, Actually, InAll, ToConclude ... \\
		\bottomrule
	\end{tabular}
 	\caption{Categories of function symbols. `SA' and `NA' are short for
		\emph{Sufficient Assumption} and \emph{Necessary Assumption} 
		respectively.}
        \label{functions}
\end{table}

\paragraph{Implementation} We use over 100 pre-defined function symbols, grouped into four categories, and apply hand-crafted rules to match them in the sentence. The function symbols along with the punctuation can be divided into one or two parts, as instantiated variables. This strategy is relatively complete, illustrated in Appendix~\ref{appendix:extraction_process}.

\section{Methods}
\begin{figure*}[t]
	\large
	\centering
	\includegraphics[scale=0.48]{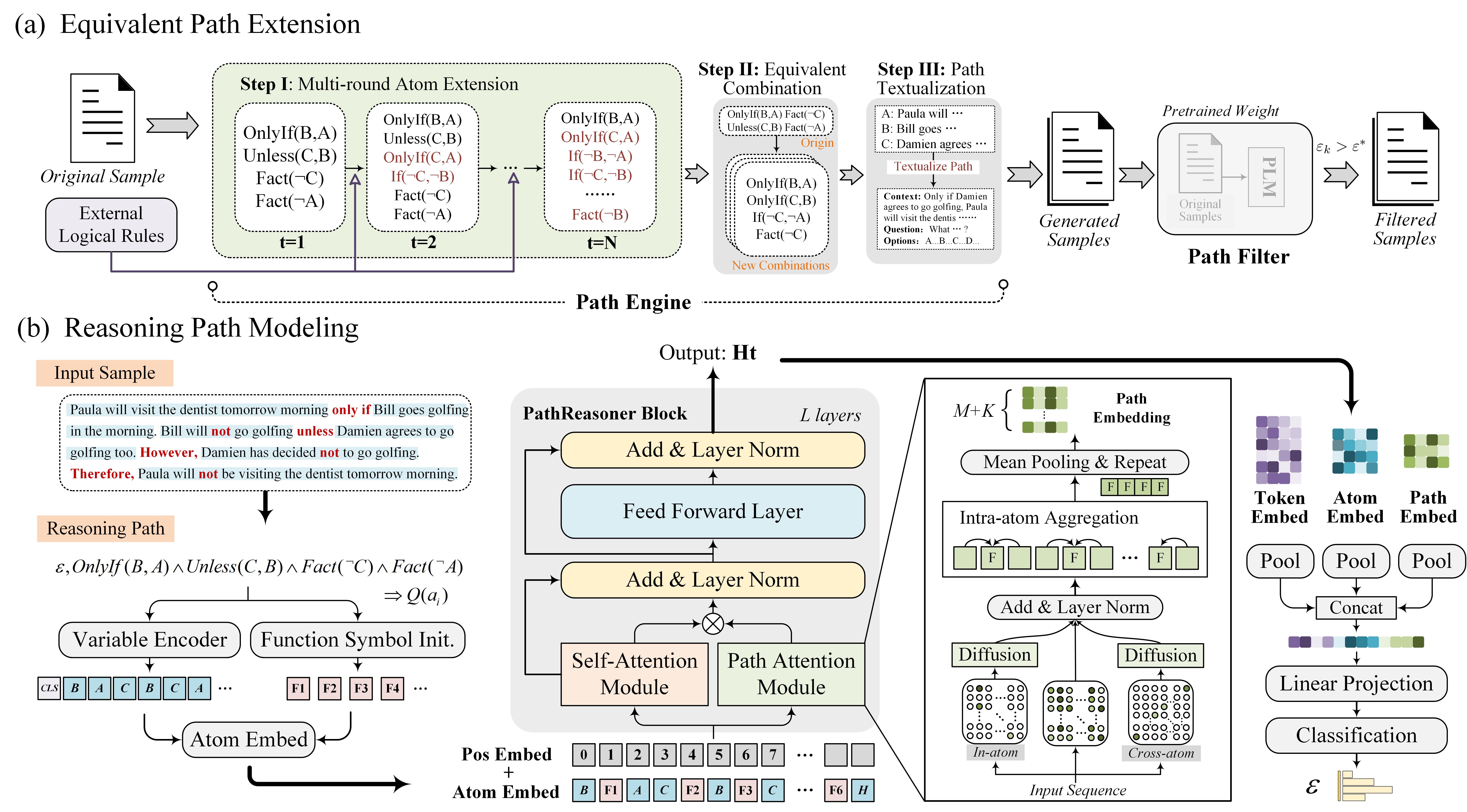}
	\caption{The architecture of PathReasoner. Part (a) is Equivalent 
		Path Extension, which aims to improve the diversity of samples. Part 
		(b) is Reasoning Path Modeling, which is designed to model logical 
		structures.}
	\label{fig_model}
\end{figure*}

To tackle the challenges in the logical reasoning task, we propose the 
architecture PathReasoner, shown in Figure \ref{fig_model}. It includes two 
main parts: (a) Equivalent Path Extension (EPE) and (b) 
Reasoning Path Modeling (RPM). The former module is aimed at expanding the 
sample diversity to improve the consistency of model prediction. The latter one 
targets at improving the logic perception capability of the reasoning model.

\subsection{Equivalent Path Extension}
After unifying the inputs into the logical rule form, it is natural to exploit the equivalent logic to facilitate the equivalent extension. 

\subsubsection{External Logical Formulas}
In the beginning, we introduce external logical formulas to achieve the atom 
extension. Corresponding to the function symbols, we employ the following 
logical formulas.

\noindent \emph{(A) Equivalence Logic.} It defines the bi-directional 
derivation between atoms as Eq. \ref{equal_logic}, where $\square \in 
\{\textrm{Cause}, 
\textrm{SA}\}$ and $\neg$ denotes the negation.

\begin{equation}
\square(A,B) \Leftrightarrow \square(\neg B, \neg A).
\label{equal_logic}
\end{equation}

\noindent \emph{(B) Single Atom Derivation}. Such logical formula is targeted at 
transforming NA atoms to SA, i.e.,

\begin{equation}
\textrm{NA}(A,B) \Rightarrow \textrm{SA}(\neg A, \neg B).
\label{single_derive}
\end{equation}

\noindent \emph{(C) Multiple Atom Derivation}. Depending on the conjunction of 
atoms, we can generate more diverse text. We only present the logical formulas 
with two atom conjunction in Eq. \ref{multiple_derive_1} and 
\ref{multiple_derive_2}, since more complex situations can be derived by 
repeating the extension process.

\begin{equation}
\star(A,B) \land \bigtriangleup(B,C) \Rightarrow \star(A, C),
\label{multiple_derive_1}
\end{equation}

\begin{equation}
\textrm{Fact}(A) \land \bigtriangledown(A,B) \Rightarrow \textrm{Fact}(B).
\label{multiple_derive_2}
\end{equation}

In above equations, $\star \in \{\textrm{Cause}, \textrm{NA} ,\textrm{SA}\},$ $
\bigtriangleup \in \{\textrm{Cause}, \textrm{NA} ,\textrm{SA}\}$ and 
$\bigtriangledown \in \{\textrm{Cause}, \textrm{NA} ,\textrm{SA}\}$.

\subsubsection{Reasoning Path Engine and Filter}
Taking original reasoning paths and equivalent logical formulas as inputs, the 
reasoning path engine module aims to generate the candidate samples.

Firstly, we conduct multi-round atom extension. For example in Fig.~\ref{fig_model}(a), there exist four atoms in the original reasoning path. At 
the first round, the atom $\texttt{Unless}(C, B)$ can derive $\texttt{If}(\neg 
C, \neg B)$, and also a new atom $\texttt{OnlyIf}(C,A)$ can be added into the 
atom base through the conjunction derivation of $\texttt{Unless}(C,B)$ and 
$\texttt{OnlyIf}(B,A)$. We repeat the extension process to include all 
potential atoms. Thus, an extended atom base is formed.

Secondly, our purpose is to mine atom combinations to form new reasoning paths. 
By enumerating all possible combinations, we select the ones which can recover 
the original path in reverse. For example, the combination of 
$\texttt{OnlyIf}(B,A)$, $\texttt{OnlyIf}(C,B)$, $\texttt{Fact}(\neg C)$ and 
$\texttt{If}(\neg C, \neg A)$ is a valid candidate because it can derive the 
original path with external logical formulas.

Thirdly, we replace the variables with the corresponding text and textualize 
the reasoning path form into regular sample form (with context, question and 
options).

To reduce noise (e.g., incorrect syntax) in the newly generated candidates, we 
introduce the path filter module. Specifically, we leverage the PLM (e.g., 
RoBERTa \cite{liu2019roberta}) to train the original samples from the 
downstream datasets. Therefore, a set of weight parameters is obtained, which 
is defined as the pre-trained filter in this paper. 

When feeding each sample into the pre-trained filter, we can obtain the 
confidence score $\varepsilon_{i}$ of the $i^{th}$ reasoning path related to 
option $a_{i}$. The predicted option ${a_k}$ is derived with the maximum 
confidence scores. We keep the samples with both correct predictions and high 
scores, which means ${a_k} = {a^ * }$ and ${\varepsilon _k} > {\varepsilon ^ * 
}$. $a_{k}$ is the predicted option with confidence score $\varepsilon_{k}$. 
$a^*$ is 
the ground-truth option and $\varepsilon^*$ is the threshold that controls 
the effectiveness of the reasoning path filter.

\subsection{Reasoning Path Modeling}
From the model view, we propose the reasoning path modeling module.
Given the input context, question, and options of one sample, we first unify 
them into the form of the reasoning path based on~\ref{preliminary}. The 
initial representation of instantiated variable set $V = \{V_1,V_2,...,V_K\}$ 
and function symbols set $S = \{S_1,S_2,...,S_M\}$ can be acquired respectively, 
where $K$ and $M$ are the number of variables and function symbols in the 
sample.

For the variable $V_{k}$ with token sequence 
$\{v_1^{(k)},v_2^{(k)},...,v_{|V_k|}^{(k)}\}$, we leverage the LM as the 
encoder to obtain the token-level embedding $\{\textbf{v}_1^{(k)}, 
\textbf{v}_2^{(k)}, ..., \textbf{v}_{|V_k|}^{(k)}\}$. Thus, its initial 
representation $\textbf{V}_k \in \mathbb{R}^{d}$ is calculated by the average 
pooling. We randomly initialize the representations for the function symbol 
$S_{m}$:

\begin{equation}
\textbf{V}_k = \frac{1}{|V_k|}\sum\limits_{i = 1}^{V_k} 
{\mathbf{v}_i^{(k)}}, \quad
\textbf{S}_m = \texttt{Init}(S_m).
\end{equation}

By aligning the variables and function symbol for each atom, we can form the 
atom embedding sequence $\textbf{A} \in \mathbb{R}^{(M+K) \times d}$. To take 
the order feature into consideration, we include the position 
embedding~\cite{DBLP:conf/nips/VaswaniSPUJGKP17} to the input sequence:

\begin{equation}
\mathbf{A}_i = \mathbf{A}_i + \texttt{PosEmbed}(A_i),
\end{equation}
where $\mathbf{A}_i$ is the embedding of the $i^{th}$ unit in $\mathbf{A}$, 
which can be either a variable or a function symbol. In this way, PathReasoner 
implements the sequential representation of logical rules.

To perform message passing over the reasoning paths, we propose a stack of 
$L$ layer blocks in a similar style of Transformer. Specifically, we feed the 
input sequence into both the self-attention and the proposed path attention 
module.

For the self-attention module of the $l^{th}$ layer, we follow the regular 
method, which projects the input sequence into query $\mathbf{Q}^{(l)} \in 
\mathbb{R}^{(M+K)\times d}$, key $\mathbf{K}^{(l)} \in \mathbb{R}^{(M+K)\times 
	d}$ and value $\mathbf{V}^{(l)} \in \mathbb{R}^{(M+K)\times d}$ by the 
projection matrices. Then, the output of the self-attention module can be 
derived as $\mathbf{H}^{(l)}_{SA}$.

For simplicity, we omit the description of multi-head attention in the main 
paper, but the selection of head number will be discussed in Appendix~\ref{appendix_hyper}.

For the path attention module, we first obtain the interaction matrix 
$\mathbf{M}^{(l)}_{seq} \in \mathbb{R}^{(M+K) \times (M+K)}$ by self 
multiplication of the input sequence $\mathbf{A}$. It models the interaction 
between any two units. Besides, the importance of each unit can be 
further considered from the perspective of in-atom and cross-atom. 

In-atom interaction models the information aggregation within one atom. Take 
the atom $S_i(V_{j}, V_{k})$ with two variables $V_{j}, V_{k}$ and one 
function symbol $S_{i}$ as an example ($i$,$j$ and $k$ are index in the input 
sequence), the attention score can be computed as:

\begin{equation}
s_{in}^{(l)} = {\rm LeakyReLU}(\mathbf{W}_{in}^{(l)} {\rm 
	tanh}(\mathbf{V}^{(l)} || \mathbf{S}_i^{(l)})),
\end{equation}
where $\mathbf{S}_i^{(l)} \in \mathbb{R}^{d}$ denotes the embedding of function 
symbol $S_i$. $\mathbf{V}^{(l)} \in \mathbb{R}^{d}$ is obtained by averaging 
the variable embedding $\mathbf{V}_{j}^{(l)}$ and $\mathbf{V}_{k}^{(l)}$. For 
atom with a single variable, the average step can be omitted. $||$ represents the 
concatenation between feature vectors. $\mathbf{W}$ is the trainable projection parameters 
(the same below).

To embed the in-atom attention, we leverage a score matrix 
$\mathbf{M}_{in}^{(l)} \in \mathbb{R}^{(M+K)\times(M+K)}$:

\begin{equation}
\mathbf{M}_{in}^{(l)} (i,j) = \mathbf{M}_{in}^{(l)} (i,k) =
\left\{ \begin{array}{l}
s_{in}^{(l)}, S_i(V_j,V_k) \text{ exists}\\
-\infty, \text{otherwise}
\end{array}. \right.
\end{equation}
We define $\mathbf{M}_{in}^{(l)}$ as a symmetric attention matrix, 
thus there also exist $\mathbf{M}_{in}^{(l)} (j,i)=\mathbf{M}_{in}^{(l)} (i,j)$ 
and $\mathbf{M}_{in}^{(l)} (k,i)=\mathbf{M}_{in}^{(l)} (i,k)$.

Cross-atom interaction models the message passing over different atoms. For 
the same variable $V_p$ and $V_q$ ($p$, $q$ are unit index of the input 
sequence), the attention score is obtained:

\begin{equation}
s_{crs}^{(l)} = {\rm LeakyReLU}(\mathbf{W}_{crs}^{(l)}((\mathbf{V}_p^{(l)} +
\mathbf{V}_q^{(l)})/2)),
\end{equation}
where $\mathbf{V}_{p}^{(l)}$ and $\mathbf{V}_{q}^{(l)}$ are the embeddings of 
two instantiated variables.

Similar to in-atom attention, we obtain a cross-atom score matrix 
$\mathbf{M}_{crs}^{(l)} \in \mathbb{R}^{(M+K) \times(M+K)}$:

\begin{equation}
\mathbf{M}_{crs}^{(l)} (p,q) =
\left\{ \begin{array}{l}
s_{crs}^{(l)}, \text{if } V_p, V_q \text{ co-occurs} \\
-\infty, \text{otherwise}
\end{array} \right.
\end{equation}

Since these two attention matrices only model one-order interaction between 
related units, the long-distance message passing is limited. Also, we extract 
the atom based on the explicit logical connectives, it ignores the implicit 
interactions within the logical text. Therefore, we introduce a 
diffusion aggregation strategy~\cite{DBLP:conf/nips/ZhaoDDKT21,liu2021deep} to achieve high-order attention:

\begin{equation}
\mathbf{M}_{in-h}^{(l)} = \sum\limits_{i = 1}^N \alpha_i 
(\mathbf{M}_{in}^{(l)})^{i},
\end{equation}

\begin{equation}
\mathbf{M}_{crs-h}^{(l)} = \sum\limits_{i = 1}^N \beta_i 
(\mathbf{M}_{crs}^{(l)})^{i},
\end{equation}
where $N$ is the maximum order number, $\alpha_i$ and $\beta_i$ are the 
trade-off coefficients to control 
the diffusion procedure. In this way, the one-order attention flow can be 
efficiently diffused to high-order relations. We can update the feature 
of sequence $\mathbf{H}_{seq}^{(l)} \in \mathbb{R}^{(M+K)\times d}$ through 
joint utilization of these three attention matrices:

\begin{equation}
\mathbf{H}_{seq}^{(l)} = {\rm softmax}(\mathbf{M}_{seq}^{(l)} + 
\mathbf{M}_{in-h}^{(l)} + \mathbf{M}_{crs-h}^{(l)})  \mathbf{A}.
\end{equation}

Within each atom, we aggregate the instantiated variable embedding in the 
function symbol to acquire a sequence of atom embedding, which can be represented as 
$\{\mathbf{H}_{S_1}^{(l)},...,\mathbf{H}_{S_M}^{(l)} \}$. 
Next, we define the reasoning path $\mathbf{H}_{p}^{(l)} \in \mathbb{R}^{d}$ 
embedding as:

\begin{equation}
\mathbf{H}_{p}^{(l)} = {\rm MeanPool} (\mathop {||}\limits_{i = 1}^M 
\mathbf{H}_{S_i}^{(l)}).
\end{equation}

To align the output embedding of the self-attention module, we repeat and stack the 
reasoning path embedding for $M+K$ times, obtaining the output of 
path-attention module $\mathbf{H}_{PA}^{(l)} \in \mathbb{R}^{(M+K)\times d}$. 
Note that the multi-head strategy is also applied in the path-attention module.

We obtain the optimized sequence embedding by adding $\mathbf{H}_{SA}^{(l)}$ and $\mathbf{H}_{PA}^{(l)}$. Following the common practice in the 
Transformer architecture, we feed the sequence into the feedforward block and 
obtain the final output $\mathbf{H}_{t}^{(l)}$ of $l^{th}$ layer.

After the respective mean pooling process on $\mathbf{H}_{cls}$, $\mathbf{H}_t^{(L)}$ and $\mathbf{H}_p^{(L)}$, the three features are concatenated and projected for the final prediction.

\section{Experiments}

This section provides comparison experiments with other strong baselines on two logical reasoning benchmarks. Extensive ablation studies and generalization evaluations are also followed.

\subsection{Datasets and Baselines}

The main experiments are conducted on two logical reasoning 
datasets ReClor \cite{yu2019reclor} and LogiQA \cite{liu2021logiqa}. To verify
the superiority of PathReasoner, we compare it with strong baselines, including 
RoBERTa-large \cite{liu2019roberta}, DAGN \cite{huang2021dagn}, FocalReasoner 
\cite{ouyang2021fact}, LReasoner \cite{wang2022logic}, AdaLoGN 
\cite{li2022adalogn}, MERIt \cite{jiao2022merit}, Logiformer 
\cite{xu2022logiformer}, as well as LLMs like text-davinci-003, GPT-3.5-turbo and PaLM 2. All the experiments are conducted with a single GPU of Tesla A100. All detailed experimental settings are listed in Appendix~\ref{appendix_hyper}.

\subsection{Comparison Results}

\begin{table*}[t]
	\small
	\centering
	\begin{tabular}{cp{2.8cm}|ccccc|ccc}
		\toprule
		\multicolumn{2}{c|}{\multirow{2}{*}{\textbf{Model}}} & 
		\multicolumn{5}{c|}{\textbf{ReClor}} 
		&\multicolumn{3}{c}{\textbf{LogiQA}} \\
		\multicolumn{2}{c|}{} &Valid &Test &Test-E &Test-H &$\Delta$ &Valid 
		&Test &$\Delta$\\
		\hline
		\multirow{7}{*}{\rotatebox{90}{Sequence}} & Random &25.00 &25.00 &25.00 
		&25.00 &- &25.00 &25.00 &- \\
		&Human Performance &- &63.00 &57.10 &67.20 &\cellcolor{gray!15}-1.10 &- 
		&86.00 &- \\
		&BERT-Large &53.80 &49.80 &72.00 &32.30 &\cellcolor{gray!60}-14.30 
		&34.10 &31.03 &\cellcolor{gray!60}-13.98 \\
		&XLNet-Large &62.00 &56.00 &75.70 &40.50 &\cellcolor{gray!50}-8.10 &- 
		&- &- \\
		&RoBERTa-Large &62.60 & 55.60 &75.50 &40.00 &\cellcolor{gray!50}-8.50 
		&35.02 &35.33 &\cellcolor{gray!55}-9.68 \\
		&LReasoner & 66.20 &62.40 &- &- &\cellcolor{gray!15}-1.70 &38.10 &40.60 
		&\cellcolor{gray!30}-4.41 \\
		&MERIt \dag &\underline{69.40} & 61.60 & 79.30 &47.80 
		&\cellcolor{gray!20}-2.50 
		&39.50 &42.40 &\cellcolor{gray!20}-2.61 \\
		\hline
		\multirow{4}{*}{\rotatebox{90}{Graph}} &DAGN &65.80 & 58.30 &75.91 
		&44.46 &\cellcolor{gray!35}-5.80 &36.87 &39.32 
		&\cellcolor{gray!35}-5.69 \\
		&FocalReasoner & 66.80 &58.90 &77.05 &44.64 &\cellcolor{gray!35}-5.20 
		&41.01 &40.25 &\cellcolor{gray!30}-4.76 \\
		&AdaLoGN &65.20 &60.20 &\underline{79.32} &45.18 
		&\cellcolor{gray!25}-3.90 &39.94 &40.71 &\cellcolor{gray!30}-4.30 \\
		&Logiformer & 68.40 & \underline{63.50} &79.09 &\textbf{51.25} 
		&\cellcolor{gray!10}-0.60  &\underline{42.24} 
		&\underline{42.55} &\cellcolor{gray!20}-2.46 \\
		\hline
		\multirow{4}{*}{\rotatebox{90}{LLM}} &text-davinci-003$^\clubsuit$ &53.00 &- &- &- &- &- &41.00 &- \\
		&GPT-3.5-turbo$^\clubsuit$ &58.80 &- &- &- &- &- &40.25 &- \\
            &GPT-4-0125-preview &84.40 &- &- &- &- &- &58.37 \\
		&PaLMv2$^\clubsuit$ &56.00 &- &- &- &- &- &48.00 &- \\
		\hline
		& PathReasoner &\textbf{70.40} &\textbf{64.10} &\textbf{80.91} 
		&\underline{50.89} &- &\textbf{43.16} &\textbf{45.01} &-\\ 
		\bottomrule
		
	\end{tabular}
 	\caption{Experimental results on ReClor and LogiQA. The percentage 
		signs (\%) of accuracy values are omitted. The optimal and sub-optimal 
		results are marked in bold and underlined (comparisons do not include LLMs). The column $\Delta$ presents the improvements of 
		PathReasoner on the test split. $\dag$ means the utilization of extra data. $\clubsuit$ denotes results from \cite{xu2023large}.}
	\label{tab:RECLOR_main}
\end{table*}

The results of comparison experiments are presented 
in Table \ref{tab:RECLOR_main}. Compared with previous SOTA baselines, 
PathReasoner presents superiority.

In ReClor dataset, PathReasoner outperforms all the graph-based methods. Compared with the SOTA method Logiformer, PathReasoner achieves improvements of 2.00\% and 0.60\% on the validation and test splits respectively. PathReasoner also shows superiority over all sequence-based methods, especially outperforming MERIt by 2.50\% on the test split. Importantly, it surpasses 
human performance, i.e., 64.10\% vs 63.00\%, which greatly pushes the boundary 
of machine reasoning. In LogiQA dataset, PathReasoner still shows competitive 
performances, improving the SOTA results by 2.46\% on the test split. PathReasoner demonstrates excellent performance and generalization in logical reasoning, as evidenced by its consistent results across two benchmarks.

Compared with representative LLMs, PathReasoner exhibits great superiority with a wide margin in the ReClor dataset. Also in the LogiQA dataset, it outperforms both text-davinci-003 and GPT-3.5 with obvious advantages and only falls behind PaLM 2 which is over x1000 in size.
 
\subsection{Ablation Studies}

\begin{table}[t]
	\small
	\centering
	\begin{tabular}{p{2.72cm}|cc|cc}
		\toprule
		\multirow{2}{*}{\textbf{Model}} & 
		\multicolumn{2}{c|}{\textbf{ReClor}} 
		&\multicolumn{2}{c}{\textbf{LogiQA}} \\
		&Valid &Test &Valid &Test\\
		\hline
		PathReasoner &70.40 &64.10 &43.16 &45.01 \\
		\hline
		\emph{EPE Part} &&&& \\
		\quad \emph{w/o whole} &67.00 &60.40 &41.16 &43.01 \\
		\quad\quad\quad $\Delta$ &\cellcolor{gray!25}-3.40 
		&\cellcolor{gray!25}-3.70 &\cellcolor{gray!20}-2.00 
		&\cellcolor{gray!20}-2.00 \\
		\quad \emph{w/o path filter} &68.40 &62.80 &42.70 &43.78 \\
		\quad\quad\quad $\Delta$ &\cellcolor{gray!20}-2.00 
		&\cellcolor{gray!15}-1.30 &\cellcolor{gray!10}-0.46 
		&\cellcolor{gray!15}-1.23 \\
		\hline
		\emph{RPM Part} &&&& \\
		\quad \emph{w/o whole} &63.00 &56.20 &38.40 &39.17 \\
		\quad\quad\quad $\Delta$ &\cellcolor{gray!45}-7.40 
		&\cellcolor{gray!45}-7.90 &\cellcolor{gray!30}-4.76 
		&\cellcolor{gray!35}-5.84 \\
		\quad \emph{w/o path attention} &67.60 &60.80 &41.94 &43.16 \\
		\quad\quad\quad $\Delta$ &\cellcolor{gray!20}-2.80 
		&\cellcolor{gray!25}-3.30 &\cellcolor{gray!15}-1.22 
		&\cellcolor{gray!15}-1.85 \\
		\quad \emph{w/o in-atom att.} &70.00 &62.80 &42.09 &44.85 \\
		\quad\quad\quad $\Delta$ &\cellcolor{gray!10}-0.40 
		&\cellcolor{gray!15}-1.30 &\cellcolor{gray!15}-1.07 
		&\cellcolor{gray!10}-0.16 \\
		\quad \emph{w/o cross-atom att.} &67.80 &62.40 &43.63 &42.70 \\
		\quad\quad\quad $\Delta$ &\cellcolor{gray!20}-2.60 
		&\cellcolor{gray!15}-1.70 &\cellcolor{gray!5}+0.47 
		&\cellcolor{gray!20}-2.31 \\
		\quad \emph{w/o diffusion} &69.00 &61.80 &42.70 &43.63 \\
		\quad\quad\quad $\Delta$ &\cellcolor{gray!15}-1.40 
		&\cellcolor{gray!20}-2.30 &\cellcolor{gray!10}-0.46 
		&\cellcolor{gray!15}-1.38 \\
		
		\bottomrule
	\end{tabular}
        \caption{Ablation studies on ReClor and LogiQA.}
	\label{tab:ablation}
\end{table}

Ablation studies for two main parts EPE and RPM in Table \ref{tab:ablation}. 
For \emph{w/o whole} of EPE, we remove the whole part of EPE and only utilize 
the origin samples for training. The performance witnesses obvious drops of 
3.70\% and 2.00\% on the two datasets respectively. For \emph{w/o path filter}, 
we keep all the new paths to generate samples without filtering. The results 
prove the effectiveness of it with 1.30\% and 1.23\% gains on the test 
respectively. For \emph{w/o whole} of RPM, we ablate the whole RPM and simply 
leverage the input sequence to predict the answer through a text encoder and a 
classifier. In this case, the model degenerates to RoBERTa-large baseline with 
more samples from EPE part. The results prove that the modeling of path 
significantly enhances the reasoning process.

To deeply verify modules in RPM, we carry out the following ablation studies. 
For \emph{w/o path attention}, we remove the path attention module. The 
performance gains prove that it is key to RPM part. For \emph{in-atom att.} and 
\emph{cross-atom att.}, we respectively ablate the attention modeling within 
atoms and across atoms. The former benefits the ReClor dataset a lot, while 
the latter is more helpful to the LogiQA dataset. It illustrates that in-atom 
and cross-atom attention are complementary to each other. For \emph{w/o 
	diffusion}, we remove the high-order diffusion strategy. Experiments show 
that the diffusion strategy is also vital to RPM part.

\begin{figure}[t]
	\begin{minipage}{\linewidth}
		\large
		\centering
		\includegraphics[scale=0.36]{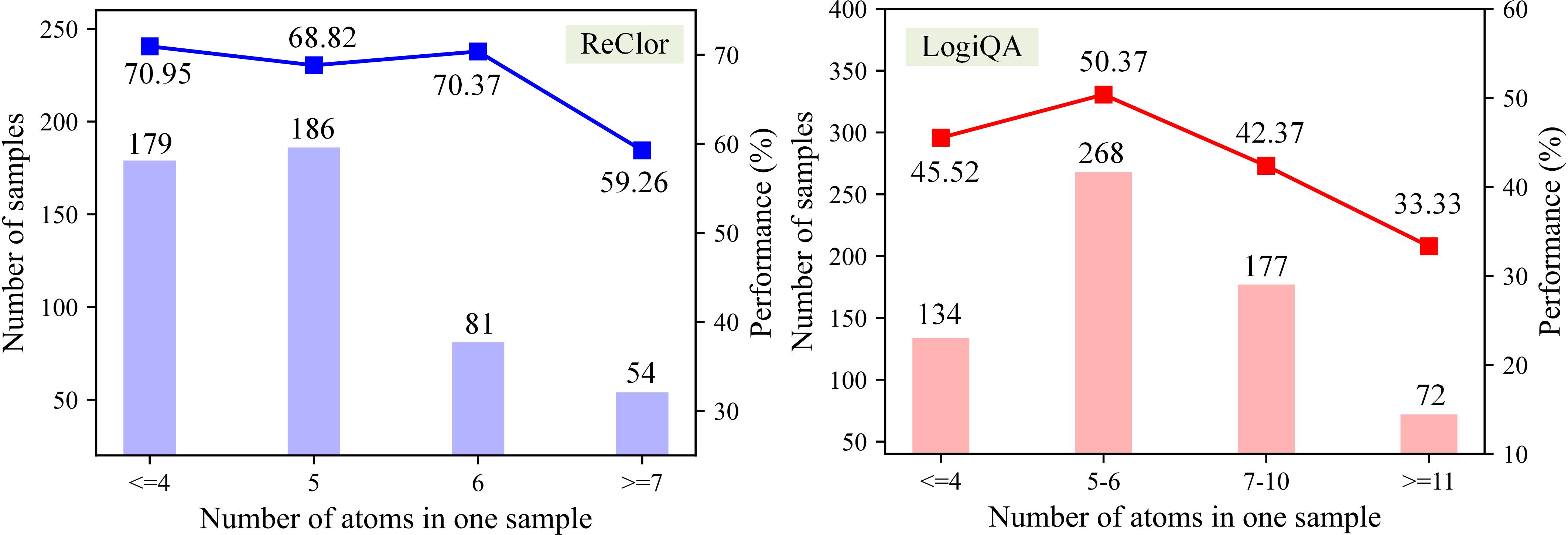}
		\subcaption{Performances with different numbers of atoms.}
		\label{atom}
	\end{minipage}
	
	\begin{minipage}{\linewidth}
		\large
		\centering
		\includegraphics[scale=0.37]{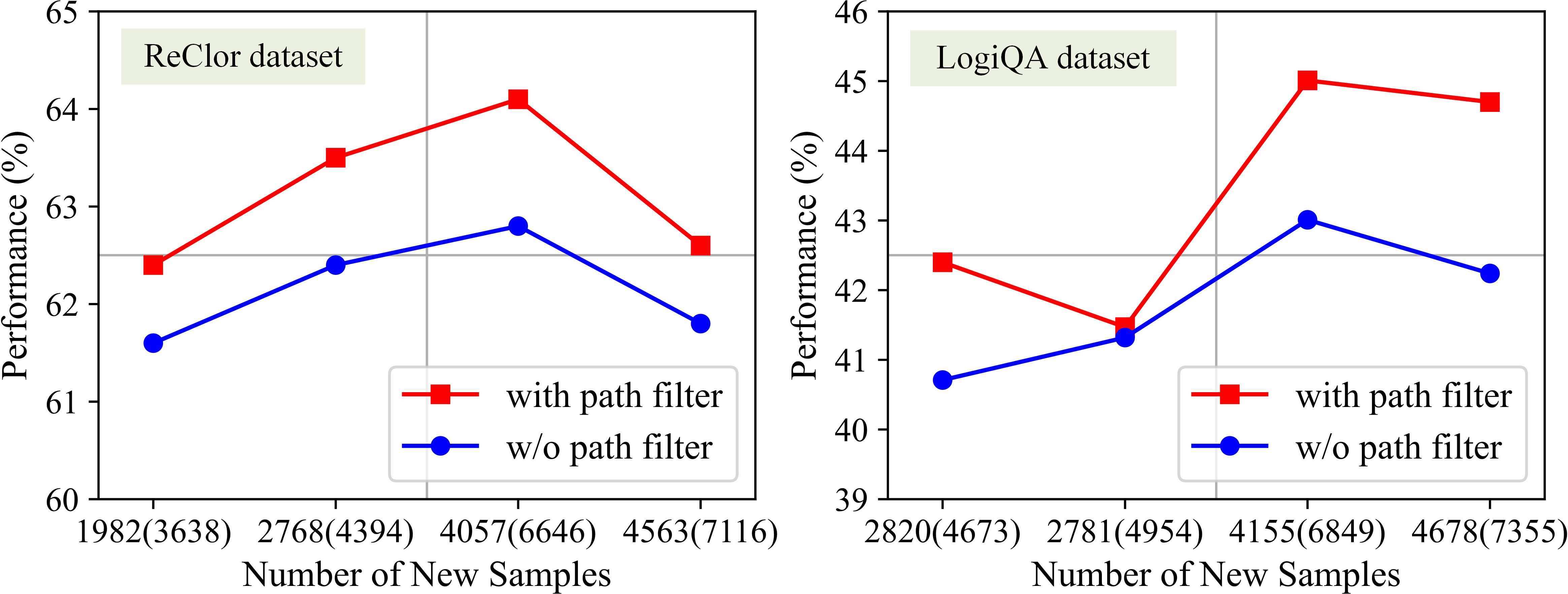}
		\subcaption{Performances with different numbers of new samples.}
		\label{number}
	\end{minipage}
	
	\begin{minipage}{\linewidth}
		\large
		\centering
		\begin{minipage}{0.48\linewidth}
			\large
			\centering
			\includegraphics[scale=0.35]{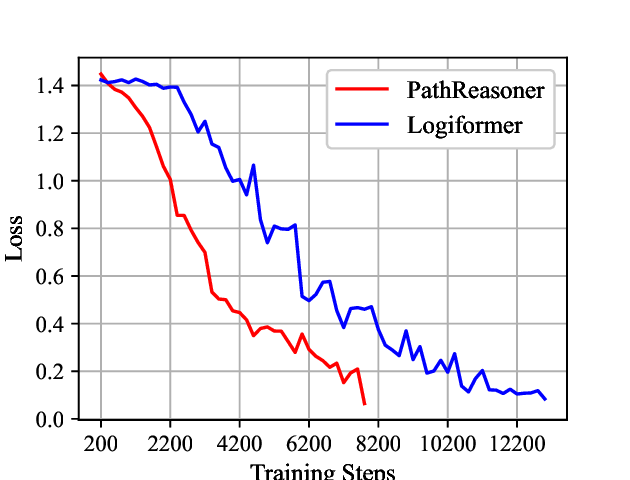}
		\end{minipage}
		\begin{minipage}{0.48\linewidth}
			\large
			\centering
			\includegraphics[scale=0.35]{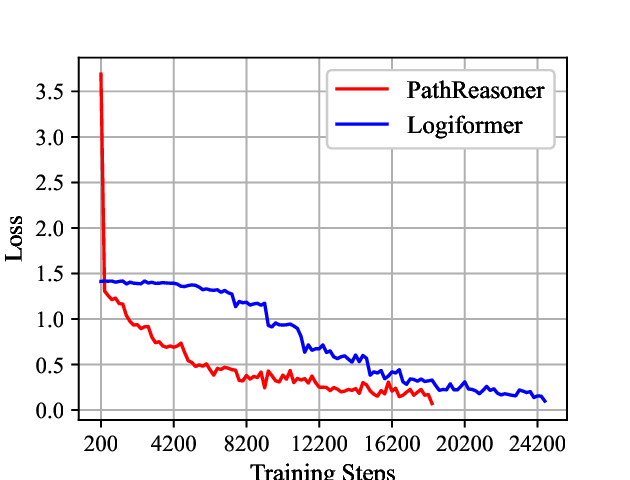}
		\end{minipage}
		\subcaption{Training Efficiency Analysis.}
		\label{training}
	\end{minipage}
	\caption{In-depth analysis of the model.}
\end{figure}

\subsection{In-depth Analysis}

We first analyze the model performances with different lengths of atoms in Fig.~\ref{atom}. The bars represent the number of samples with different atom 
numbers, while the lines denote the performances with different atom numbers. For both ReClor and LogiQA datasets, PathReasoner maintains a high 
performance with moderate scale of atoms, which accounts for most 
samples in both datasets. Confronted with larger sample sizes, the performances decline. We argue that the gaps have been 
greatly narrowed with the proposed diffusion strategies, compared with previous 
models.

Secondly, we provide an analysis of the impact of the number of new samples. 
By controlling the maximum scale of new atom combinations, we can generate 
different numbers of samples. Fig.~\ref{number} shows the model performances 
under various cases, where the horizontal axis denotes the number of new samples 
(with \& w/o path filter) and the vertical axis is the model performance on the 
test. On the two datasets, the path filter plays a positive role in reducing 
redundancy and noise. Additionally, the optimal results are obtained at a 
moderate scale of new samples, and larger amounts of samples do not always bring 
gains in performance.

Thirdly, we discuss the model training efficiency in Fig.~\ref{training}. We 
make the comparison with the previous SOTA Logiformer on ReClor (left) and LogiQA 
(right). To make a clear illustration, we report the loss curve with steps (truncated at 0.1). From the results, PathReasoner shows 
faster convergence speed on both ReClor and LogiQA datasets. Detailedly, 
PathReasoner achieves 1.66x convergence speed than Logiformer on the ReClor 
dataset, and it has 1.34x speed on the LogiQA dataset. We provide more in-depth 
experiments in Appendix \ref{in_depth}.

\subsection{Model Generalization}
PathReasoner is also evaluated on other reasoning tasks to verify the 
generalization capability in Table \ref{tab:generalize}.
The experiments are 
conducted on Dream \cite{sun-etal-2019-dream} and MuTual 
\cite{cui-etal-2020-mutual}, which are multi-turn dialogue datasets requiring 
complex reasoning. We utilize RoBERTa-Large model and the previous SOTA model 
Logiformer as baselines. Among all comparison metrics, PathReasoner 
achieves consistent superiority over them. Compared with 
Logiformer, PathReasoner outperforms it with 3.08\% in the test split of Dream, 
0.89\% of the R@1 metric of MuTual and 1.81\% of the R@1 metric of MuTual$^+$. 
It demonstrates that PathReasoner can well generalize to different reasoning 
tasks. 
Also, other generalization experiments on EPE module and zero-shot settings are included in Appendix \ref{appendix:generalization_epe},\ref{zero-shot}.

\begin{table}[t]
	\small
	\centering
	\begin{tabular}{p{1.7cm}|cc|c|c}
		\toprule
		\multirow{2}{*}{\textbf{Model}} & 
		\multicolumn{2}{c|}{\textbf{Dream}} 
		&\textbf{MuTual} &\textbf{MuTual}$^+$ \\
		&Valid &Test &R@1 &R@1 \\
		\hline
		RoBERTa-L &83.18 &84.74 &87.46 &80.47 \\
		Logiformer &84.47 &83.76 &88.04 &79.68 \\
		\hline
		PathReasoner &\textbf{85.05} &\textbf{86.84} &\textbf{88.93} 
		&\textbf{81.49} \\
		\bottomrule
	\end{tabular}
        \caption{Experiments on model generalization.}
	\label{tab:generalize}
\end{table}

\begin{figure}[t]
	\large
	\centering
	\includegraphics[scale=0.49]{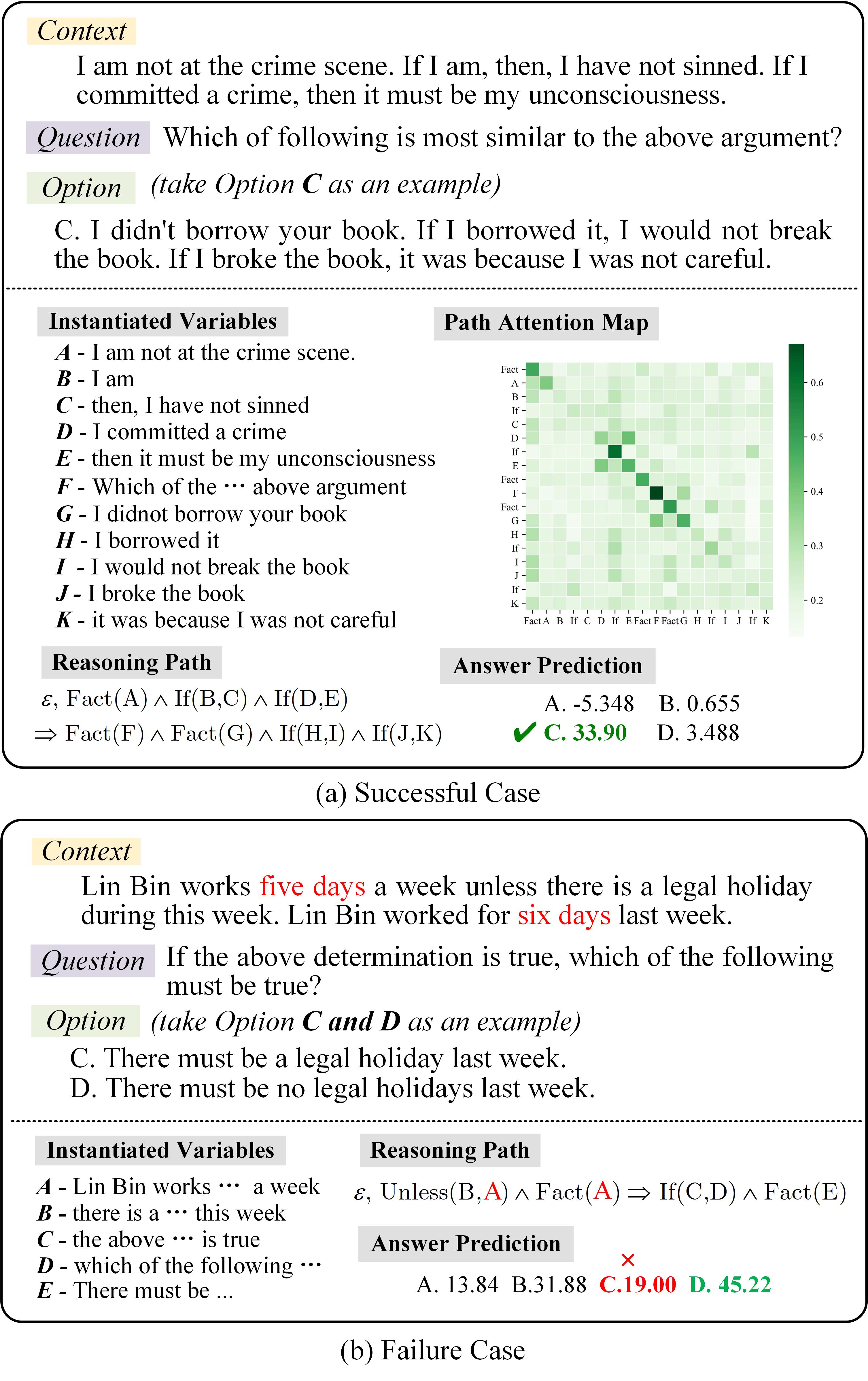}
	\caption{Two case studies on LogiQA dataset.}
	\label{casestudy}
\end{figure}

\subsection{Case Study}
We provide the analysis for the interpretability of PathReasoner in Figure \ref{casestudy}. In the successful case, PathReasoner 
correctly extracts the variables from the text and forms the reasoning path. 
In particular, We present the path attention map from RPM part to check the logical 
perception capability. Firstly, PathReasoner focuses more on the function 
symbols (e.g., $\texttt{If}$ and $\texttt{Fact}$) and question sentences (i.e., 
variable $F$), with higher attention scores in the map. It verifies that 
PathReasoner is equipped with the perception of logic and question types. 
Secondly, the question is to match the logical structure between context and 
option. The corresponding atoms (e.g., $\texttt{If}(D,E)$ and 
$\texttt{If}(J,K)$) are considered together in the module. It illustrates that 
PathReasoner is good at understanding the question and reasoning over paths.

In the failure case, PathReasoner wrongly categorizes the variable with 
different semantics together to $A$ which leads to the mistake. It demonstrates 
that the variable extraction in PathReasoner is not good at distinguishing the 
minor difference, which has space for improvement.

\section{Conclusion}
To tackle the logical data scarcity and weak model perception of logical 
structures, we propose a new paradigm to model the logical reasoning 
task by representing each natural sentence as atom form and transforming 
logical samples into reasoning paths. Based on such unique modeling, an 
architecture PathReasoner is proposed to address the challenges.
It achieves SOTA performances on two logical reasoning 
datasets. Also, extensive experiments demonstrate the effectiveness of each 
module and great generalization capability on other complex reasoning 
scenarios.
In the future, we will propose a unified architecture based on PathReasoner to tackle the logical reasoning tasks over different modalities (e.g., images, text, graphs). 

\section{Acknowledgement}
This work was supported by National Key Research and Development Program of China (2022YFC3303600), National Natural Science Foundation of China (62137002, 62293550, 62293553, 62293554, 61937001, 62250066, 62176209, 62176207, and 62192781), "LENOVO-XJTU" Intelligent Industry Joint Laboratory Project, Natural Science Basic Research Program of Shaanxi (2023-JC-YB-593), the Youth Innovation Team of Shaanxi Universities, XJTU Teaching Reform Research Project "Acquisition Learning Based on Knowledge Forest".

\section*{Limitations}
This paper proposes a novel direction for addressing logical reasoning tasks, which differs from the sequence-based methods and graph-based methods.
The core of the proposed model is to transform the input text into the form of logical rules with the conjunction of atoms and realize the equivalent extension and path reasoning over it. However, the extraction process of atoms is still very challenging. Although the current algorithm predefines some basic logical relations in advance and achieves great progress, it also requires the help of more comprehensive external logic bases in the future to improve the accuracy of atom extraction. In addition, the logical text in reality often contains noise (e.g., wrong logic). Although this paper has conducted extensive experiments on other reasoning datasets and complex settings to verify the generalization capability, there still remain unsolved on how to promote the models to more complex settings, like multi-modality scenarios.

\bibliography{anthology,custom}
\bibliographystyle{acl_natbib}

\clearpage
\appendix

\section{Pilot Experiments}
\label{pilot}
In this section, we provide the detailed pilot experiments mentioned in Fig. 1 of the main paper. For the model prediction consistency test, we equally replace the explicit logical connectives in a part of the samples on ReClor. The differences in performances are presented in Table \ref{pilot1}.

\begin{table}[h]
	\small
	\centering
	\caption{Pilot experiments on prediction consistency.}
	\begin{tabular}{p{1.7cm}|ccc}
		\toprule
		\textbf{Model} &\textbf{Origin} &\textbf{Replace}&$\Delta$\\
		\hline
		BERT-L &38.50 &30.00 &-8.50 \\
		RoBERTa-L &55.00 &48.50 &-6.50 \\
		\hline
		PathReasoner &62.50 &61.00 &-1.50 \\
		\bottomrule
		
	\end{tabular}
	\label{pilot1}
\end{table}

It can be seen that current PLMs fail to maintain equal predictions on samples 
with the same logical semantics. It proves the motivation of the proposed method. Also, we provide the performances of PathReasoner in the same setting as the pilot experiments. Our model largely improves the prediction consistency, and only fails in 1.50\% of the cases. It illustrates the robustness of PathReasoner in logic.

In addition, we conducted experiments on the model perception of logical connectives. By adding, deleting, or modifying the explicit logical connectives on some samples, we randomly break the original semantics of the context. We report the ratio of samples that fail to follow the logical changes. It tests the 
sensitivity of the model for capturing the logical relations. Results are shown in Table \ref{pilot2}.

\begin{table}[h]
	\small
	\centering
	\caption{Pilot experiments on model perception of explicit logical connectives.}
	\begin{tabular}{p{1.9cm}|c}
		\toprule
		\textbf{Model} &\textbf{Ratio}\\
		\hline
		BERT-L & 29.30\% \\
		RoBERTa-L & 21.63\% \\
		\hline
		PathReasoner &71.95\% \\
		\bottomrule
		
	\end{tabular}
	\label{pilot2}
\end{table}

From the results, current PLMs are not always sensitive to the changes of logical connectives. BERT and RoBERTa can merely distinguish 29.30\% and 21.63\% of changes respectively. Therefore, it is worth considering enhancing the logic modeling for the language models, which supports our motivations. Also, we report the performance of PathReasoner on the last row of the table. Our model shows great superiority on enhancing the model perception of explicit logical connectives, being sensitive to 71.95\% of the cases. It well verifies and supports our motivations. 

\section{Key Questions for Extraction Process}
\label{appendix:extraction_process}
The whole extraction process leads to several key questions:

\paragraph{(1) Scenario coverage.} Our predefined rules are relatively complete, and have covered extensive cases in syntax (guided by experts). We include over 100 instantiated function symbols (curated from NLTK) and it can cover most logical scenarios (details in Appendix~\ref{appendix:statistics}). Therefore, it can ensure the wide coverage of logical scenarios. Beyond that, we also include the \emph{Fact} category of function symbols. It can be adapted to facts that do not have obvious logic. To sum up, our heuristic rules can extend to any kind of text in theory.

For an example out of the logical domain, the input is factual paragraph X, which consists of sentences A, B, C, and D. Our method adapts to such a scenario, and it can output $Fact(A) \land Fact(B) \land Fact(C) \land Fact(D)$. 

\paragraph{(2) Extraction accuracy.} Based on the above descriptions, our method can cover any kind of text in theory. We randomly select 30 paragraphs, resulting in 148 pieces of sentences. We manually label the extraction accuracy of each sentence. To make a comparison, we also prompt GPT-4 (instruction+predefined function symbols + atom form+4-shot examples) to finish this process. The results are listed in Table~\ref{exp:extraction_acc}.

\begin{table}[h]
\small
\centering
\caption{Experiments on the extraction accuracy.}
\begin{tabular}{p{1.3cm}ccc}
    \toprule
    &\textbf{Ours} &\textbf{GPT-4} &\textbf{LLaMA-2-Chat} \\
    \hline
    Atom Acc &95.27 &91.22 &8.11 \\
    \bottomrule
\end{tabular}
\label{exp:extraction_acc}
\end{table}

\section{Distinction of Our Logical Forms}
\label{appendix:distinction_logic}

As some examples presented, our predefined logical forms are similar to first-order logic (FOL) and propositional logic. But our forms are more suitable for the scenarios in the following aspects.

\paragraph{(1) Customized function symbols.} We define four types of function symbols and they are effective in equivalent transformation. The general FOL and propositional logic can not satisfy our customized requirements.

\paragraph{(2) Perception of sentence-level logic.} In logical reasoning scenarios, rich logic exists more at the sentence level, thus we transform each sentence into an atom. However, FOL and propositional logic are conditioned at the entity level or span level, which is more fine-grained. They are not necessarily effective in capturing logic.

\section{Statistics of Function Symbols}
\label{appendix:statistics}
In this section, we present the statistics of the logical connectives in two 
logical reasoning datasets ReClor and LogiQA. It will provide intuitive 
proof of the necessity and rationality of the function symbol categories. 

\begin{figure}[t]
	\begin{minipage}{\linewidth}
		\large
		\centering
		\includegraphics[scale=0.45]{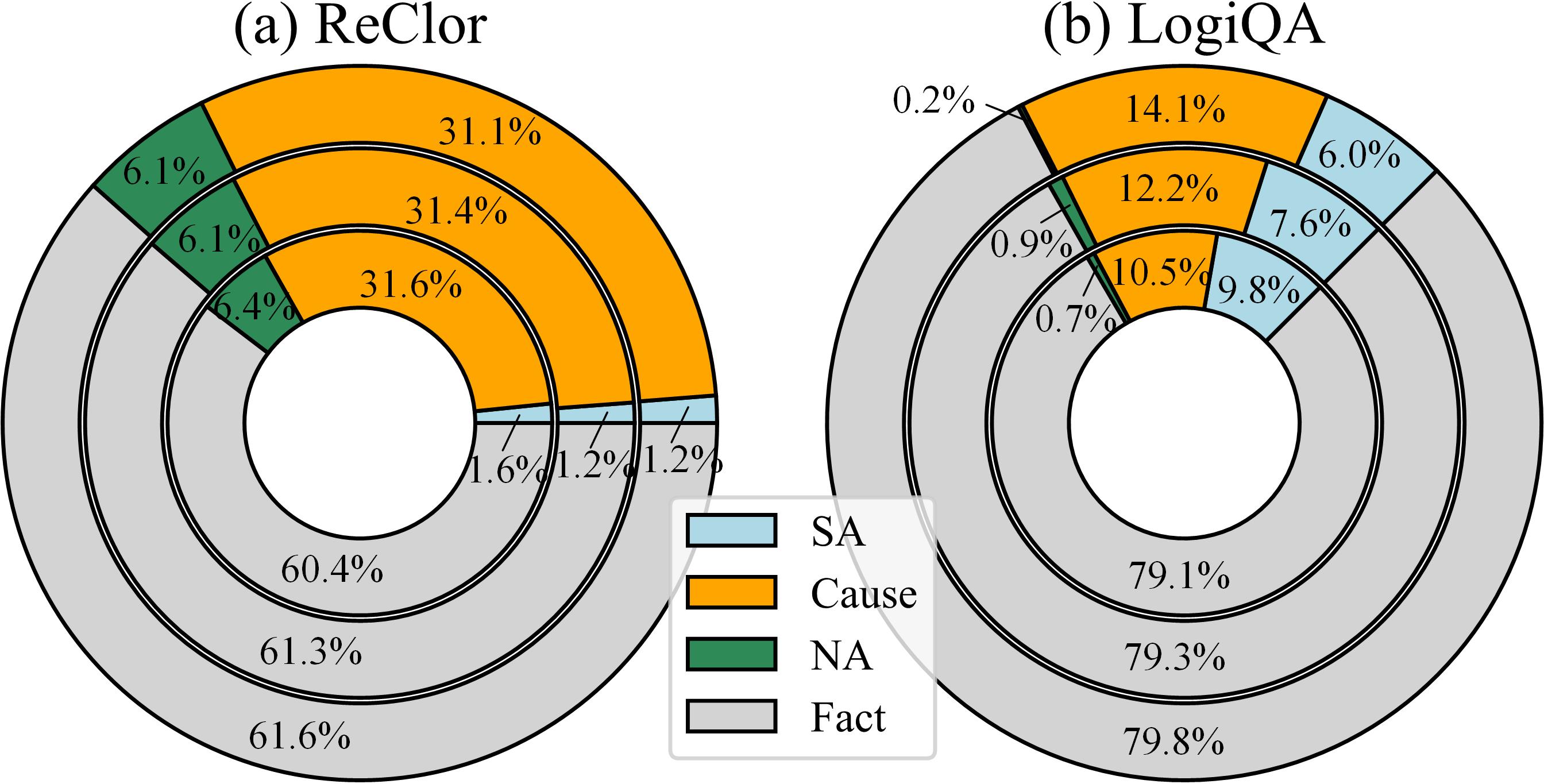}
		\subcaption{Statistics of function symbols in train (outer cycle), 
			validation (middle cycle) and test (inner cycle) splits.}
		\label{statistics1}
	\end{minipage}
	
	\begin{minipage}{\linewidth}
		\large
		\centering
		\includegraphics[scale=0.19]{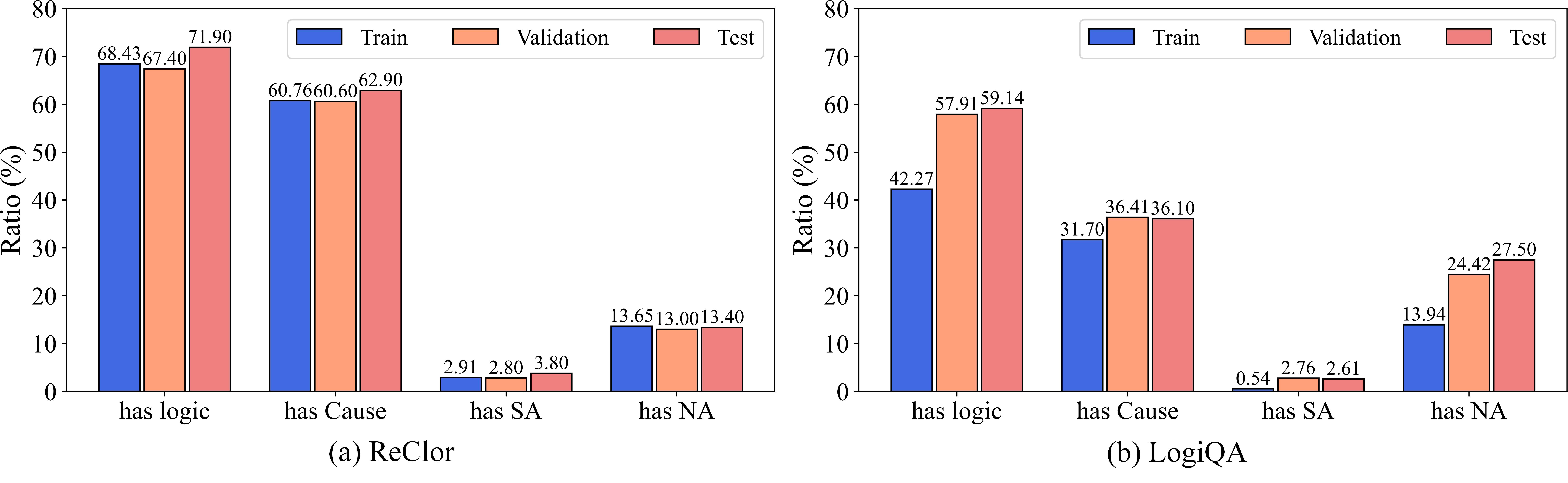}
		\subcaption{Statistics of logical samples.}
		\label{statistics2}
	\end{minipage}
	\caption{Statistics of logical reasoning benchmarks.}
\end{figure}

Figure \ref{statistics1} presents the statistics of function symbols (i.e., 
\emph{Cause}, \emph{SA}, \emph{NA}, \emph{Fact}) in the context of two 
benchmarks. The outer cycle represents the train split, the middle one is the 
validation split and the inner one is the test split. In the ReClor dataset, 
nearly 40\% of atoms are non-fact, which contain explicit logical connectives 
(i.e., \emph{Cause}, \emph{SA}, \emph{NA}). Among them, \emph{Cause} relations 
are the majority. In the LogiQA dataset, the ratio of the logical function symbols 
drops a lot, but it still accounts for about 20\%. 

Figure \ref{statistics2} shows the statistics of logical samples. We categorize 
the samples with any one of the three logical function symbols into \emph{has 
	logic}. Similar, we include samples with \emph{Cause}, \emph{SA} and \emph{NA} 
to \emph{has Cause}, \emph{has SA} and \emph{has NA} respectively. In ReClor, 
nearly 70\% of the samples have explicit logical connectives. Also, over 60\% of 
samples contain \emph{Cause} atoms. In LogiQA, samples with logical connectives 
account for 50\%. The ratio of samples containing \emph{Cause} atoms drops to 
about 35\% while the ratio of samples with \emph{NA} atoms increases.

The above analysis illustrates that the two benchmark datasets are abundant in 
logical connectives. Thus, the modeling of logical atoms is of great necessity.

\section{Experimental Settings}
\label{detail_set}
\subsection{Benchmarks and other Datasets}
\label{appendix_datasets}
ReClor and LogiQA are two representative datasets for the logical reasoning 
task. The details are presented as follows.

\noindent \textbf{ReClor} \cite{yu2019reclor} includes 6,138 samples total with 4,638 training 
samples, 500 validation samples, and 1,000 samples for test. All of them are 
collected from some standardized graduate admission examinations. To 
discriminate the difficulty of the questions, the test split is divided into 
$\emph{Test-E}$ and $\emph{Test-H}$, where the former represents the easy 
version of the test samples and the latter denotes the harder parts.

\noindent \textbf{LogiQA} \cite{liu2021logiqa} includes 8,678 samples sourced 
from National Civil 
Servants Examinations of China. It is further split into the training set, 
development set, and test set, with 7,376, 651, and 651 samples respectively. 

Also, to verify the model generalization capability, we employ two dialogue 
datasets involving complex reasoning, which are Dream and MuTual. Also, we 
exploit the zero-shot logical reasoning capability of the proposed model on the 
recently proposed ZsLR benchmark. The details are presented below.

\noindent \textbf{Dream} \cite{sun-etal-2019-dream} contains 6,444 multiple 
choice questions, sourced from English-as-a-foreign-language examinations. The 
samples are split into train, development and test sets with 3,869, 1,288 and 
1,287 samples respectively. We report the exact match metric on both validation 
and test splits.

\noindent \textbf{MuTual} \cite{cui-etal-2020-mutual} consists of 8,860 
questions, divided into 7,088 training samples, 886 validation samples, 886 
test samples. It is modified from Chinese high school English listening 
comprehension test data. Also, MuTual$^{plus}$ dataset is proposed to test 
whether the model is capable of selecting a safe response when necessary. Since 
the test split of MuTual is not made public, we only report the R@1 metric 
(recall at position one) on the validation set.

\noindent \textbf{ZsLR} \cite{xu2023mind} includes 6 zero-shot splits modified 
from ReClor dataset. Since the dataset contains 17 reasoning types in total, 
some types of samples are classified as seen types during training. For the test, 
it defines two metrics, one is $\emph{Test-All}$ which tests on all the types 
of samples, and another is $\emph{Test-Unseen}$ which only tests on the unseen 
parts of types.

\begin{table}[h]
	\centering
        \footnotesize
	\caption{Categorization of recent works on logical reasoning task. `DA' 
		denotes the data augmentation strategy. `$\dag$' denotes the 
		utilization of 
		extra data.}
	\scalebox{0.99}{
	\begin{tabular}{p{1.8cm}|ccc|c}
		\toprule
		\textbf{Model} &\textbf{Sequence} &\textbf{Graph} &\textbf{Path/Rule} 
		&\textbf{DA}\\
		\hline
		ReClor &\checkmark & & \\
		DAGN  & &\checkmark & \\
		FocalReasoner  & &\checkmark & \\
		LReasoner &\checkmark & & &\checkmark \\
		AdaLoGN & &\checkmark & & \\
		MERIt \dag &\checkmark & & &\checkmark \\
		Logiformer & &\checkmark & &\\
		\hline
		PathReasoner & & &\checkmark &\checkmark \\ 
		\bottomrule
	\end{tabular}}
	\label{tab:categorize}
\end{table}

\begin{table*}[h]
    \centering
        \footnotesize
    \caption{The details of tuned hyper-parameters on the two logical reasoning 
        benchmarks.}
    \begin{tabular}{p{5.3cm}|p{2.9cm}<{\centering}c|p{2.9cm}<{\centering}c}
        \toprule
        \multirow{2}{*}{\textbf{Name of Parameter}} & 
        \multicolumn{2}{c|}{\textbf{ReClor}} 
        &\multicolumn{2}{c}{\textbf{LogiQA}} \\
        &Search Scope &Best &Search Scope &Best \\
        \hline
        \emph{General Settings} & & & &\\
        \quad number of epoch &\{10,12,15,20\} &20 &\{10,12,15,20\} &20 \\
        \quad max sequence length &\{384,512\} &384 &\{384,512\} &512 \\
        \quad learning rate &\{4e-6, 5e-6, 6e-6\} &5e-6 &\{4e-6, 
        5e-6, 6e-6\} &5e-6 \\
        \hline
        \emph{Equivalent Path Extension} & & & &\\
        \quad path filter threshold $\varepsilon^*$ &\{0.5,0.8,0.9\} &0.9 
        &\{0.5,0.8,0.9\} &0.9\\
        \hline
        \emph{Reasoning Path Modeling} & & & &\\
        \quad number of layer &\{3,4,5,6\} &3 &\{3,4,5,6\} &3 \\
        \quad number of head &\{4,8\} &4 &\{4,8\} &4 \\
        \quad max diffusion order $N$ & \{1,2,3\} &2 &\{1,2,3\} &2\\
        \quad in-atom diffusion $\alpha_1$ 
        &\{0,0.1,0.2,0.3,0.4\} &0.2 &\{0,0.1,0.2,0.3,0.4\} &0.1 \\
        \quad cross-atom diffusion $\beta_1$ 
        &\{0,0.1,0.2,0.3,0.4\} &0 &\{0,0.1,0.2,0.3,0.4\} &0.1 \\
        \quad leaky rate &\{0.01,0,02,0.03,0.04\} &0.02 
        &\{0.01,0,02,0.03,0.04\} &0.02 \\
        \bottomrule
    \end{tabular}
    \label{hyper}
\end{table*}

\subsection{Baselines}
\label{appendix_baselines}
In this paper, we compare PathReasoner with all the previous methods of the 
logical reasoning task, including the SOTA model Logiformer. There methods can 
be categorized into sequence-based and graph-based, shown in Table 
\ref{tab:categorize}.

\noindent \textbf{(1) Random}. The results are obtained from the random 
predictions.

\noindent \textbf{(2) RoBERTa-Large} \cite{liu2019roberta}. The trained 
language model RoBERTa is employed as the text encoder to obtain the 
predictions. It is also the same with the baselines of BERT-Large 
\cite{kenton2019bert} and XLNet-Large \cite{yang2019xlnet}.

\noindent \textbf{(3) Human Performance} \cite{yu2019reclor,liu2021logiqa}. The 
performances are averaged from the scores of some graduate students on the test 
split. 

\noindent \textbf{(4) DAGN} \cite{huang2021dagn}. It is the first graph-based 
work to tackle the logical reasoning task. It splits the text into nodes and 
leverages the graph neural networks to reason over the chain-type graph.

\noindent \textbf{(5) FocalReasoner} \cite{ouyang2021fact}. It focuses on the 
facts within the context and it extracts all the fact units to form a 
supergraph for reasoning.

\noindent \textbf{(6) LReasoner} \cite{wang2022logic}. It proposes to leverage 
the defined rules (e.g., De Morgan's Laws) to extend the context. In addition, 
it employs data augmentation strategies (e.g., contrastive learning) to 
improve the diversity of the samples.

\noindent \textbf{(7) MERIt} \cite{jiao2022merit}. It proposes a meta-path 
guided strategy to conduct the pretraining on the external corpus. The 
pre-trained module is further verified based on some off-the-shelf SOTA 
methods. For a fair comparison, we directly derive the results of MERIt with the RoBERa-Large backbones from the original paper.

\noindent \textbf{(8) AdaLoGN} \cite{li2022adalogn}. It first builds a text 
graph based on the off-the-shelf method and models it in an adaptive 
neuro-symbolic system.

\noindent \textbf{(9) Logiformer} \cite{xu2022logiformer}. It models the 
context from the perspective of both logic and syntax, building a causal graph 
and a co-occurrence graph. Specifically, it reasons on the graph transformer 
networks with biased attention.

Additionally, we include the following representative large language models to make the comparisons.

\noindent \textbf{(10) text-davinci-003}. It was created by OpenAI, of which the training data was collected up to Sep. 2021. The size of text-davinci-003 is 175B.

\noindent \textbf{(11) GPT-3.5-turbo}. It is also from OpenAI and the training corpus is collected up to June. 2021. GPT-3.5-turbo is of the same size as text-davinci-003.

\noindent \textbf{(12) PaLM 2}. It was created by Google. It has a larger size than the above two LLMs, which is 540B.

The results of the three LLMs on the logical reasoning benchmarks are collected from \cite{xu2023large}.

\subsection{Implementation Details}
\label{appendix_hyper}
In the implementation, to make a fair comparison, we employ the RoBERTa-large 
\cite{liu2019roberta} model with 
the hidden size of 1024 as the encoder of text. We utilize the Adam 
\cite{kingma2014adam} for the optimization. Also, we set different 
hyper-parameters for the two logical reasoning datasets respectively. We tune 
some of the hyper-parameters for the optimal within a scope. Table \ref{hyper} 
presents the detailed information.

The listed hyper-parameters belong to three parts: general settings, equivalent 
path extension module, and reasoning path modeling module. Considering the 
calculation cost, we do not utilize the grid search strategy, instead, we 
sequentially search the hyper-parameters for the optimal. For the reasoning 
path modeling module, we select the maximum diffusion order $N$ to be 2. 
Therefore, there only exist four diffusion trade-off co-efficient $\alpha_1$, 
$\alpha_2$, $\beta_1$, $\beta_2$, which satisfy $\alpha_1+\alpha_2=1$ and 
$\beta_1+\beta_2=1$. So we only list the tuning details of in-atom diffusion 
trade-off $\alpha_1$ and cross-atom diffusion trade-off $\beta_1$.

\section{In depth Analysis}
\label{in_depth}
In this section, we provide more experiments to analyze the model performances.

\subsection{Model Performance on Multiple Logics}
In the category of function symbols, we take \emph{Cause}, \emph{SA}, \emph{NA} 
and \emph{Fact} into consideration. Among them, the first three represent the 
logical relations (non-fact) while the last one represents the factual 
expression. Therefore, we give an analysis of how our model performs on these 
factual or logical samples. We test the model performance on three types of 
samples: (1) Fact, where all atoms are factual; (2) Simple Logic, where there 
only exists one category of logical function symbols in each sample; (3) 
Complex Logic, where multiple categories of logical function symbols are 
included in one sample. Table \ref{complex} presents the results of 
PathReasoner on the above settings, compared with RoBERTa-Large model and 
Logiformer. Since ReClor does not make the test split public, we only report 
the results on the validation split.

\begin{table}[h]
	\small
	\centering
	\caption{Experiments on multiple logics on ReClor.}
	\begin{tabular}{p{1.7cm}|c|c|c}
		\toprule
		\textbf{Model} &\textbf{Factual} &\textbf{Simple} &\textbf{Complex}  \\
		\hline
		RoBERTa-L &67.65&65.56&56.79 \\
		Logiformer &67.65&71.48&63.58 \\
		\hline
		PathReasoner &\textbf{72.06}&\textbf{73.33}&\textbf{64.81} \\
		\bottomrule
		
	\end{tabular} 
	\label{complex}
\end{table}

For factual types of samples, PathReasoner achieves 4.41\% gains over the 
baselines. We argue that previous method like Logiformer focuses too much on 
the capture of logical relations but fails to better generalize to the 
fact-only samples. PathReasoner leverages the atom form to represent both the 
logical content and the factual content, thus it can also improve the 
performances on factual samples. For simple logic samples and complex logic 
samples, PathReasoner also shows the superiority of 1.85\% and 1.23\% over 
Logiformer respectively. It demonstrates the competitiveness of PathReasoner in 
logical perception and reasoning. Meanwhile, we witness that PathReasoner 
does excellent in capturing simple logic and maintaining factual 
reasoning, but there still exists space for improvement on the complex logic.

\begin{table*}[h]
	\small
	\centering
	\caption{The details of ReClor Test Split on different reasoning types. 
		\textbf{NA}: Necessary Assumption, \textbf{S}:Strengthen, 
		\textbf{W}:Weaken, \textbf{E}:Evaluation, \textbf{I}:Implication, 
		\textbf{ER}:Explain or Resolve, \textbf{T}:Technique, \textbf{IF}:Identify 
		a Flaw, \textbf{MF}:Match Flaws, \textbf{MS}:Match the Structure, 
		\textbf{O}:Others.}
	\begin{tabular}{p{1.8cm}|cccccccccccccc}
		\toprule
		\textbf{Model} &\textbf{NA} &\textbf{S} &\textbf{W} &\textbf{E} 
		&\textbf{I} &\textbf{ER} &\textbf{T} &\textbf{IF} &\textbf{MF} 
		&\textbf{MS} &\textbf{O}\\
		\hline
		PathReasoner &74.56 
		&62.77&59.29&76.92&52.17&67.86&83.33&67.52&58.06&83.33&67.12 \\
		\hline
		Logiformer 
		&74.56&64.89&55.75&76.92&45.65&61.90&66.67&58.12&45.16&66.67&60.27\\ 
		\quad\quad $\Delta$ 
		&-&\cellcolor{green!15}-2.12&\cellcolor{red!15}+3.54&-&\cellcolor{red!15}+6.52&\cellcolor{red!15}+5.96&\cellcolor{red!15}+6.66
		&\cellcolor{red!15}+9.40 &\cellcolor{red!15}+12.90 
		&\cellcolor{red!15}+6.66 &\cellcolor{red!15}+6.85\\
		\hline
		RoBERTa-L 
		&71.05&61.70&47.79&69.23&39.13&58.33&52.78&61.54&45.16&56.67&52.05 \\
		\quad\quad $\Delta$ &\cellcolor{red!15}+3.51 &\cellcolor{red!15}+1.07 
		&\cellcolor{red!15}+11.50 &\cellcolor{red!15}+7.69 
		&\cellcolor{red!15}+13.04 &\cellcolor{red!15}+9.53 
		&\cellcolor{red!15}+30.55 &\cellcolor{red!15}+5.98 
		&\cellcolor{red!15}+12.90 &\cellcolor{red!15}+16.66 
		&\cellcolor{red!15}+15.07\\
		\bottomrule
	\end{tabular}
	\label{types}
\end{table*}

\begin{table*}[h]
	\small
	\centering
	\caption{Experimental results on 6 zero-shot logical reasoning splits. 
		\emph{T-A} and \emph{T-U} denote the abbreviations of the metrics 
		\emph{Test-All} and \emph{Test-Unseen} respectively.}
	\begin{tabular}{p{2.2cm}|p{0.65cm}<{\centering}p{0.65cm}<{\centering}|p{0.65cm}<{\centering}p{0.65cm}<{\centering}|p{0.65cm}<{\centering}p{0.65cm}<{\centering}|p{0.65cm}<{\centering}p{0.65cm}<{\centering}|p{0.65cm}<{\centering}p{0.65cm}<{\centering}|p{0.65cm}<{\centering}p{0.65cm}<{\centering}}
		\toprule
		\multirow{2}{*}{\textbf{Model}} & \multicolumn{2}{c|}{\textbf{v1}} & 
		\multicolumn{2}{c|}{\textbf{v2}} & \multicolumn{2}{c|}{\textbf{v3}} & 
		\multicolumn{2}{c|}{\textbf{v4}} & \multicolumn{2}{c|}{\textbf{v5}} & 
		\multicolumn{2}{c}{\textbf{v6}}\\
		& T-A & T-U & T-A & T-U & T-A & T-U
		& T-A & T-U & T-A & T-U	& T-A & T-U\\
		\hline
		BERT-Large &38.00&34.36&42.00&33.39&37.50&31.61&38.00&33.26&29.60 
		&28.02&28.80&32.24 \\
		RoBERTa-Large 
		&47.70&39.47&50.60&39.90&46.10&40.58&50.40&42.45&53.00
		&43.66 &49.90 &50.92 \\
		DAGN&49.20&41.37&52.70&43.56&49.60
		&39.73&52.50 &44.51 &52.40 &42.63 &48.50 &49.15\\
		LReasoner  
		&46.90&40.60&50.20&43.49&48.40&42.76&49.20&44.12&51.90 
		&42.02 &46.30&44.93\\
		Logiformer &43.50&39.31&54.80 
		&46.30 &48.80&42.24&52.10 &44.85
		&52.10 &40.88 &51.50 &51.44 \\
		TaCo 
		&\underline{52.20}&\textbf{47.51}&\textbf{55.80}&\textbf{48.79} 
		&\underline{52.20}&\underline{44.26} &\underline{54.70}&\textbf{49.89}
		&\underline{56.00} &\underline{46.67} &\underline{54.70} 
		&\textbf{55.17}\\
		\hline
		PathReasoner &\textbf{52.70} &\underline{45.87} &\underline{55.10} 
		&\underline{44.01} &\textbf{52.20} &\textbf{45.43} &\textbf{56.60} 
		&\underline{49.20} &\textbf{57.20} &\textbf{47.43} &\textbf{54.90} 
		&\underline{54.28}\\
		\bottomrule
	\end{tabular}
	
	\label{tab:RECLOR_zeroshot}
\end{table*}

\begin{figure}[h]
	\begin{minipage}[t]{0.48\linewidth}
		\large
		\centering
		\includegraphics[scale=0.48]{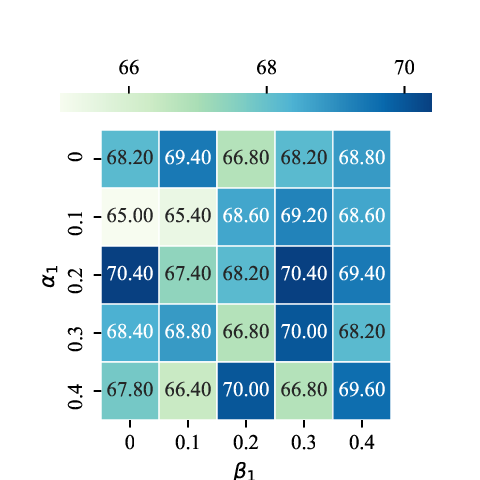}
		\subcaption{Validation split.}
		\label{diffusion_valid}
	\end{minipage}
	\begin{minipage}[t]{0.48\linewidth}
		\large
		\centering
		\includegraphics[scale=0.48]{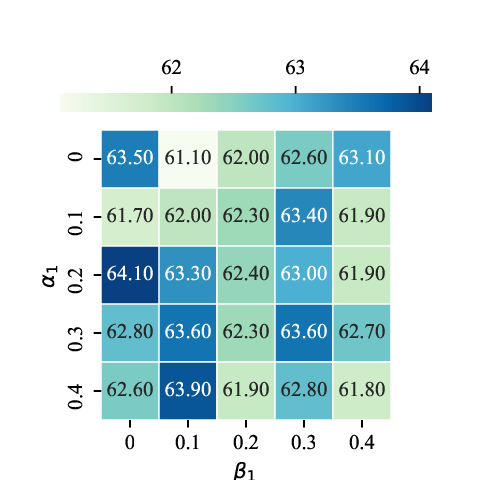}
		\subcaption{Test split.}
		\label{diffusion_test}
	\end{minipage}
	\caption{Analysis of high-order diffusion strategy.}
	\label{diffusion}
\end{figure}

\subsection{Model Performance on Different Reasoning Types}
In the ReClor dataset, the samples are divided into 17 reasoning types. Table 
\ref{types} gives in-depth model performances on different reasoning types. 
Limited by space, we only present 11 types in the table. From the results, 
PathReasoner performs better in most cases. Specially, for \emph{IF}, \emph{MF} 
and \emph{MS}, PathReasoner achieves obvious superiority. Considering that 
these reasoning types require the perception of logical structures, the gains 
in performance prove the effectiveness of PathReasoner.

\subsection{High-order Diffusion Strategy Analysis}
In the implementation, we set the maximum order of diffusion to 2, that is 
$N=2$. Therefore, we employ two trade-off coefficients $\alpha_i$ and $\beta_i$ 
to control the diffusion procedure. We search $\alpha_i$ and $\beta_i$ from the 
set of \{0, 0.1, 0.2, 0.3, 0.4\} and report the results on the test split of 
ReClor in Figure \ref{diffusion}. From the results, PathReasoner achieves the 
optimal simultaneously on both the validation and test splits.

\section{Generalization of Equivalent Path Extension Module}
\label{appendix:generalization_epe}

Beyond the main experiments on logical reasoning benchmarks, generalization experiments (Table~\ref{tab:generalize}) and zero-shot settings (Table~\ref{diffusion}), we add a simple experiment with our proposed equivalent path extension (EPE). The aim is to achieve a \textbf{plug-in-and-play} function to augment the training of LLMs. In detail, we randomly sample from the Flan collection~\cite{weifinetuned}, leading to ~80K original instruction-following samples. Then, we apply EPE to generate equivalent instruction samples. These augmented samples are leveraged to tune LLaMA-2-Chat (7B). The test experiments on MMLU (57 tasks) and BigBenchHard (21 tasks) are presented in Table~\ref{exp:gen_epe}.

\begin{table}[h]
\small
\centering
\caption{Experiments on the generalization capability of equivalent path extension module.}

\begin{tabular}{p{3.9cm}|cc}
    \toprule
    \textbf{Model} &\textbf{MMLU} &\textbf{BBH} \\
    \hline
    LLaMA-2-Chat &45.78 &35.01 \\
    LLaMA-2-Chat + Flan &46.94 &36.99 \\
    LLaMA-2-Chat + EPE + Flan &\textbf{48.75} &\textbf{38.96} \\
    \bottomrule
\end{tabular}
\label{exp:gen_epe}
\end{table}

With the EPE augmentation process, the tuned LLaMA-2-Chat can witness significant performance improvements, compared with two baselines: one is LLaMA-2-Chat, and another is LLaMA-2-Chat tuned on sampled Flan collection. Such findings largely expand the application scope of PathReasoner, especially in empowering the training of off-the-shelf LLMs.

\section{Model Generalization on Zero-shot Logical Reasoning Settings}
\label{zero-shot}
Previous work \cite{xu2023mind} argued that the ideal full-data setting is not 
sufficient to test the logical reasoning performances and has proposed a new 
benchmark for generalized zero-shot logical reasoning (named ZsLR). To verify 
the model generalization on the zero-shot settings, we conduct experiments on 
ZsLR and compare with several SOTA baselines. The results are shown in Table 
\ref{tab:RECLOR_zeroshot}.

From the results of 6 splits, PathReasoner is competitive on the majority of 
the cases compared with TaCo and Logiformer. For split v1, v3, v4, v5 and v6, 
PathReasoner outperforms all the strong baselines on the metric of 
\emph{Test-All}, which verifies the great generalization ability on both seen 
and unseen types of samples. Compared with the full-data setting SOTA model 
Logiformer, PathReasoner shows obvious superiority on all the splits and all 
the test metrics. The great advantages uncover the huge potential of modeling 
reasoning paths for the logical text, which improves the extensibility and 
generalization of the model. Also, it is worth noticing that there still exists 
space for improvement on the unseen types of samples, especially on the split 
of v2, v4, and v6.

\section{Restatement of Our Key Novelty}
\label{appendix:key_novelty}
We will clarify the obvious differences of our method compared with previous works (especially, the graph-based method). It can be divided into three points.

\paragraph{Extraction strategy.} Transforming the natural language into units is a common method in the reasoning field. However, we largely differ in the definition of relationships. Previous works only limit to a subset of relation words. For example, Logiformer only attends to causal relations as the connectives. It is sufficient for evaluations (see Appendix~\ref{appendix:statistics}), which overfit the logical reasoning benchmarks. However, our definition of function symbols is different from previous works, and our coverage is broad enough (over 100 relation words). Therefore, our method can be extended to other scenarios, which have been verified with generalization experiments and zero-shot settings.

\paragraph{Flexible extension strategy.} Benefiting from the distinctive definition of function symbols, we can formulate the context into the conjunction of atoms (i.e., reasoning paths). Therefore, we can easily conduct the equivalent path extension to derive new combinations of atoms. This advantage is distinctive from other works. It is also one of our main contributions.

In some graph-based methods, the text graph is updated to capture new relations along with the message-passing process. The whole process is extremely time-consuming, which is the main shortcoming. Our method actually decouples the dynamic extension process with the formulation of atoms and paths. It augments data diversity and improves training efficiency.

\paragraph{Incoporated advantages in the path attention module.} In fact, the path attention module combines the advantages of sequence-based and graph-based methods. Previous sequence-based methods ignore logical structures but can handle long-distance dependency with Transformer structure. Graph-based methods usually rely on GNN-style modules to update the features, but lack the extensibility to larger context and fine-grained modeling within each unit (Logiformer attempts to solve it through attention bias, but still limits to a coarse level). In our path attention module, the advantages of sequence- and graph-based methods are inherited. It further achieves differentiable and interpretable reasoning (see Case Study).

To sum up, the distinctions between PathReasoner and other methods are significant. Also, we would like to emphasize that the distinction of PathReasoner does not only benefit the logical reasoning benchmarks, which previous works are limited to. PathReasoner indeed shows strong generalization capability and plug-in-and-play property (see Appendix~\ref{appendix:generalization_epe} for details).

\end{document}